\documentclass[onecolumn]{article}

\usepackage{microtype}
\usepackage{graphicx}
\usepackage{subfigure}
\usepackage{booktabs} %
\usepackage{amsmath, bm, bbm}
\usepackage{mathrsfs}
\usepackage{amssymb}
\usepackage{float}

\usepackage{hyperref}

\newcommand{\dataset}{\mathit{dataset}}
\newcommand{\attributes}{\mathit{attributes}}
\newcommand{\ising}{\textrm{(ising)}}
\newcommand{\BIC}{\mathit{BIC}}

\usepackage[accepted]{icml2019}

\icmltitlerunning{Covering up bias with Markov blankets: A post-hoc cure for attribute prior avoidance}

\usepackage{breqn}

\begin{document}

\onecolumn
\icmltitle{Covering up bias in CelebA-like datasets with Markov blankets: \\A post-hoc cure for attribute prior avoidance}

\icmlsetsymbol{equal}{*}

\begin{icmlauthorlist}
\icmlauthor{Vinay Prabhu*}{unifyid}
\icmlauthor{Dian Ang Yap*}{stanford}
\icmlauthor{Alexander Wang}{stanford}
\icmlauthor{John Whaley}{unifyid}
\end{icmlauthorlist}

\icmlaffiliation{stanford}{Department of Computer Science, Stanford University, Stanford, California}
\icmlaffiliation{unifyid}{UnifyID, Redwood City, California}

\icmlcorrespondingauthor{Dian Ang Yap}{dayap@stanford.edu}
\icmlcorrespondingauthor{Alexander Wang}{aswang96@stanford.edu}

\icmlkeywords{Machine Learning, Flow Models, CelebA, Bias Corrections, ICML}

\vskip 0.3in

\printAffiliationsAndNotice{\icmlEqualContribution} %

\begin{abstract}
Attribute prior avoidance entails subconscious or willful non-modeling of (meta)attributes that datasets are oft born with, such as the 40 semantic facial attributes associated with the CelebA and CelebA-HQ datasets. The consequences of this infirmity, we discover, are especially stark in state-of-the-art deep generative models learned on these datasets that just model the pixel-space measurements, resulting in an inter-attribute bias-laden latent space. This viscerally manifests itself when we perform face manipulation experiments based on latent vector interpolations. In this paper, we address this and propose a post-hoc solution that utilizes an Ising attribute prior learned in the attribute space and showcase its efficacy via qualitative experiments. 
\end{abstract}

\section{Motivation: Attractive Barack Obama $\neq$ White Blonde Barack Obama}
GIF animations showcasing nifty manipulation of celebrity faces\footnote{ \url{https://youtu.be/G06dEcZ-QTg}}
by vector operations in the discovered latent space have emerged as a modality of choice in order to advertise the effectiveness of image generative models of all hues, be it GANs \cite{goodfellow2014generative}, VAEs \cite{kingma2013auto} or Flow-based models \cite{dinh2014nice}. In fact, the idea that 
deep generative models, especially flow-based generative models, can be trained without attribute labels by attempting to directly model the distribution of the input images, $\bm{x} \sim p(\bm{x})$, and yet yield latent-space representations worthy of \textit{downstream tasks} like manipulating the semantic attributes of images is seen as a marquee feature\footnote{The exact terms used are: \textit{Useful} latent space in Section~1 and \textit{meaningful} latent space in Section~6 of \cite{kingma2018glow}} of the model \cite{kingma2018glow}.
This is however tantamount to not just ignoring the rich co-occurrence structure between the image attributes in the dataset that these models are trained on but also making a faux pixel space - attribute space marginal independence assumption. 
That is, for a dataset 
$\mathcal{D} = \left\{ {\left( {{{\mathbf{x}}_i},{{\mathbf{a}}_i}} \right)} \right\}_{i = 1}^{N_{\dataset}}$ 
where $\mathbf{x}_i$ is the $i^{th}$ image and $\mathbf{a}_i$ is the attribute vector associated with that $i^{th}$ image, 
modeling just $p(\mathbf{x})$ is akin to assuming $p\left( {{\mathbf{x}},{\mathbf{a}}} \right) = p({\mathbf{x}})p({\mathbf{a}})$ given that marginalization yields: $ \sum\limits_{\mathbf{a}} {p\left( {{\mathbf{x}},{\mathbf{a}}} \right) = } \sum\limits_{\mathbf{a}} {p({\mathbf{x}})p({\mathbf{a}})}=p({\mathbf{x}}) $.

\begin{figure}[t]
\vskip 0.2in
\begin{center}
\centerline{\includegraphics[width=0.6\linewidth]{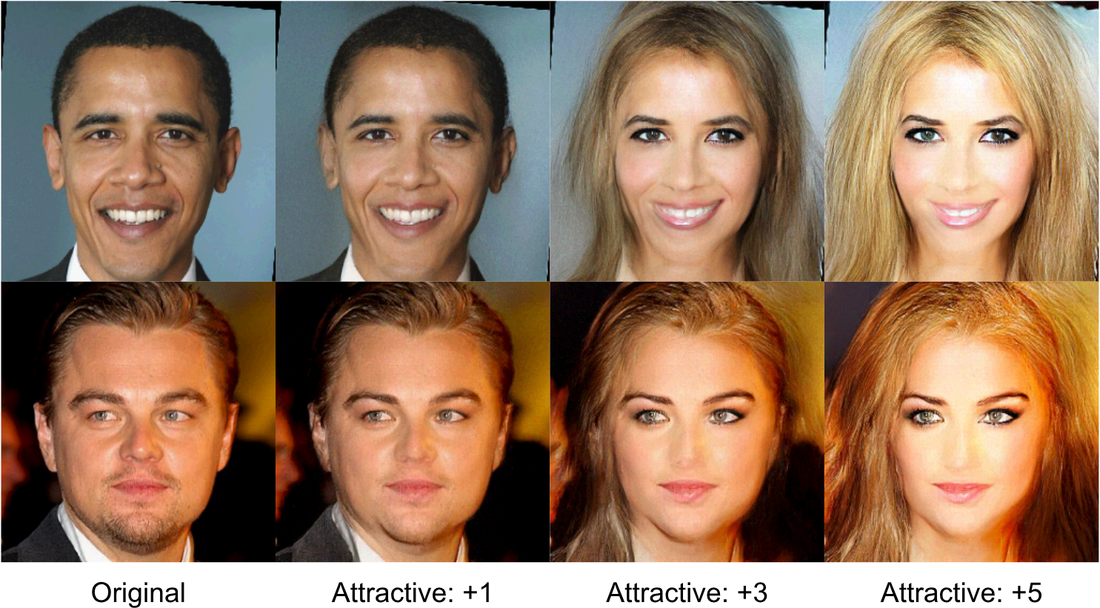}}
\caption{Interpolating along the \texttt{Attractive} latent attribute in different scales converges to pictures of white blonde women.}
\label{fig:BlondeObama}
\end{center}
\vskip -0.3in
\end{figure}

\begin{figure*}[t]
\vskip 0.2in
\begin{center}
\centerline{\includegraphics[width=0.8\linewidth]{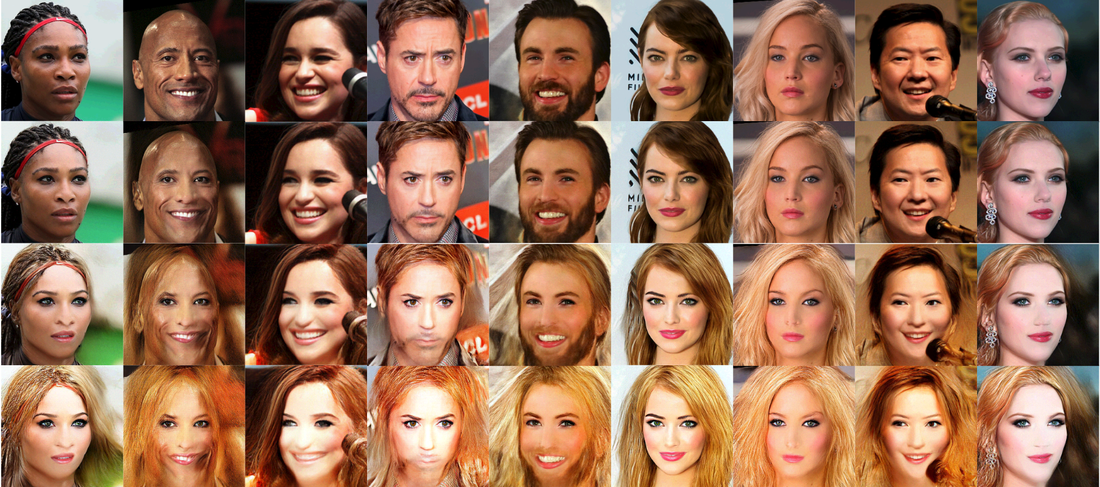}}
\caption{More visualizations of how images of different gender and races converge to young, white-skinned women with long blonde or red hair as we only naively interpolate along the 'attractive' attribute. \textit{From top to bottom: original, attractive: ($\alpha= + 1$,  attractive: $\alpha=+3$,  attractive:$\alpha= +5$.)}}
\label{fig:BlondeEveryone}
\end{center}
\vskip -0.3in
\end{figure*}

Specifically, with regards to CelebA \cite{liu2015faceattributes}, we have $N_{\dataset}=30000$ images, each with 40 binary semantic attributes. If we ignore the co-occurrence of these attributes and not account for it, we will end up with inter-attribute spill-over effects, an example of which is showcased in Figure~\ref{fig:BlondeObama}. Using the model in \cite{kingma2018glow} as an example, we see the end result of trying to \textit{enhance} the \texttt{attractive} attribute in Barack Obama's and Leonardo DiCaprio's \footnote{\url{https://www.imdb.com/name/nm0000138/}} pictures.\footnote{\url{https://openai.com/blog/glow/}} Figure~\ref{fig:BlondeEveryone} shows additional examples of varying the \texttt{attractive} attribute across a variety of images. Moreover, since CelebA-HQ was derived and cherry-picked from CelebA \cite{karras2017progressive}, the biases are also inherited and are apparent in models trained on CelebA-HQ.

As we increase the \textit{weight} of influence of the attractive-attribute manipulation latent vector, we end up rendering Obama and DiCaprio as almost Caucasian-blonde women. One can attribute this to the possible idiosyncratic co-prevalence of attributes such as \texttt{blond\_Hair}, \texttt{attractive}, and \texttt{pale\_Skin} in the images of the original dataset. 

We acknowledge this troublesome nexus\footnote{This \textit{nexus} has also been deeply explored in works such as \cite{barocasfairness} and \cite{tom2017designing}} between these three attributes and would like to reiterate that this work does not propose a \textit{`here's how you make a face more attractive'} framework but rather focuses on the issue on ignored statistical priors of the (meta)attributes that an image dataset is oft shipped with. Specifically, we address this issue by using graph structured Ising priors to model the inter-attribute statistical dependence and later harnessing this model to propose a corrective procedure that helps allay some of the concerns laid thus far.

The rest of the paper is organized as follows. In Section~\ref{learning-ising-prior}, we cover the estimation procedure for learning the Inter-Attribute Ising Prior (IAIP). In Section~\ref{methodology}, we propose our post hoc bias- corrective framework. In Section~\ref{results}, we showcase some qualitative results that highlights the efficacy of our approach and conclude the paper.

This is a work-in-progress and we have duly open-sourced the implementation along with an interactive portal provides interfaces to both the IAIP as well as the corrective procedure here\footnote{\url{https://yapdianang.github.io/celeba/}}. This work aims to showcase and correct the bias in the image domain, which draw parallels to the learnt biases in Word2Vec embeddings \cite{bolukbasi2016man} and other existing study of biases \cite{torralba2011unbiased}\cite{tommasi2017deeper}\cite{binns2017fairness}. We'd like to re-emphasize we position this work as a quick-to-implement post-hoc solution and would like to encourage the deep learning community to not just pay attention to the dubious selection-biases in the datasets they use to train their generative models, but to also consider the problem of jointly modeling the attribute space and the pixel-space of images.

\section{Learning the Inter-Attribute Ising Prior}
\label{learning-ising-prior}
Probabilistic graphical models are the unifying framework of choice in areas such as bioinformatics, statistical physics and communication theory for modeling complex dependencies among random variables using ideas from both graph theory and probability theory. This framework not just facilitates building large-scale multivariate statistical models but also allows for elegant visualization and knowledge discovery through the learned statistical dependency graph \cite{wainwright2008graphical}. In this paper, we choose a specific type of binary Markov Random Fields (MRFs) termed as \textit{Ising Models} (see \cite{stanley1971phase}) to model the inter-attribute statistical dependencies. The model specifies a distribution $p(\mathbf{a})$ defined over a \textit{substrate} graph, $G^{\ising}(V,E)$. Specifically the conditional distribution of an attribute $a_j \in \{0,1 \}$ given the rest of the attributes $a_{\backslash j}$ and the graph $G^{\ising}(V,E)$ takes the form:
\begin{dmath}
{p_\Omega }\left( {{a_j}|{a_{\backslash j}}} \right) = \frac{{\exp \left[ {{\tau _j}{a_j} + {a_j}\sum\limits_{k \in {V_{\backslash j}}} {\left( {{\beta _{jk}}{a_k}} \right)} } \right]}}{{1 + \exp \left[ {{\tau _j} + \sum\limits_{k \in {V_{\backslash j}}} {\left( {{\beta _{jk}}{a_k}} \right)} } \right]}}
\end{dmath}
Here, $\tau_j$ and $\beta_{jk}$ are the node parameters and Ising edge parameters respectively.

In order to learn the underlying graph and the associated node and edge weights, we use a network estimation procedure termed \texttt{eLasso} that combines $\ell_1$-regularized logistic regression with model selection criterion based on the Extended Bayesian Information Criterion (EBIC).\footnote{\texttt{eLASSO} is implemented in the R package \textit{IsingFit}: \url{http://cran.r-project.org/web/packages/IsingFit/IsingFit.pdf}}

The EBIC cost function is defined as: 
\begin{dmath}
\BIC_{\gamma }\left( {{{\hat \Omega }_j}} \right) =  - 2\ell \left( {{{\hat \Omega }_j}} \right) + n_{nei}(\rho)\log \left( N_{\dataset} \right) \\
+ 2\gamma n_{nei}(\rho)\log \left( {N_{\attributes} - 1} \right)
\end{dmath}
Here, $N_{\dataset}$ represents the number of
observations (which is 30,000 in the case of Celeb-A-HQ dataset), $N_{\attributes}-1$ represents the number of covariates (predictors), $n_{nei}(\rho)$ represents the cardinality of the neighbors set chosen 
by the logistic regression at a certain penalty weight $\rho$, and $\gamma$ is the \textit{prior strength} goodness-of-fit hyperparameter \cite{van2014new}. 

So, given the binary attribute matrix $A\in \{0,1\}^{N_{\dataset} \times N_{\attributes}}$, the log-likelihood cost function for the $j^{th}$ attribute, $\ell \left( {{{\hat \Omega }_j}} \right)$ is defined as:
\begin{dmath}
    \ell \left( {{{\hat \Omega }_j}} \right) = \sum\limits_{i = 1}^{N_{\dataset}} {\left( {{\tau _j}{A_{ij}} + \sum\limits_{k \in {V_{\backslash j}}} {\left( {{\beta _{jk}}{A_{ij}}{A_{ik}}} \right) - \log \left( {1 + \exp \left\{ {{\tau _j} + \sum\limits_{k \in {V_{\backslash j}}} {{\beta _{jk}}{A_{ik}}} } \right\}} \right)} } \right)} 
    \end{dmath}
After learning the node weights and the edge weights using the \texttt{IsingFit} procedure, we symmetrize the weights as:
\begin{equation}
    w_{jk}= 
\begin{cases}
    \frac{(\beta_{jk} + \beta_{kj})}{2}, & \text{if } \beta_{jk} \neq 0 \textit{ and } \beta_{kj} \neq 0\\
    0, & \text{otherwise}
\end{cases}
\end{equation}
\subsection{The learned IAIP for the CelebA-HQ dataset}

\begin{figure*}[!t]
    \centering
    \includegraphics[width=0.8\linewidth]{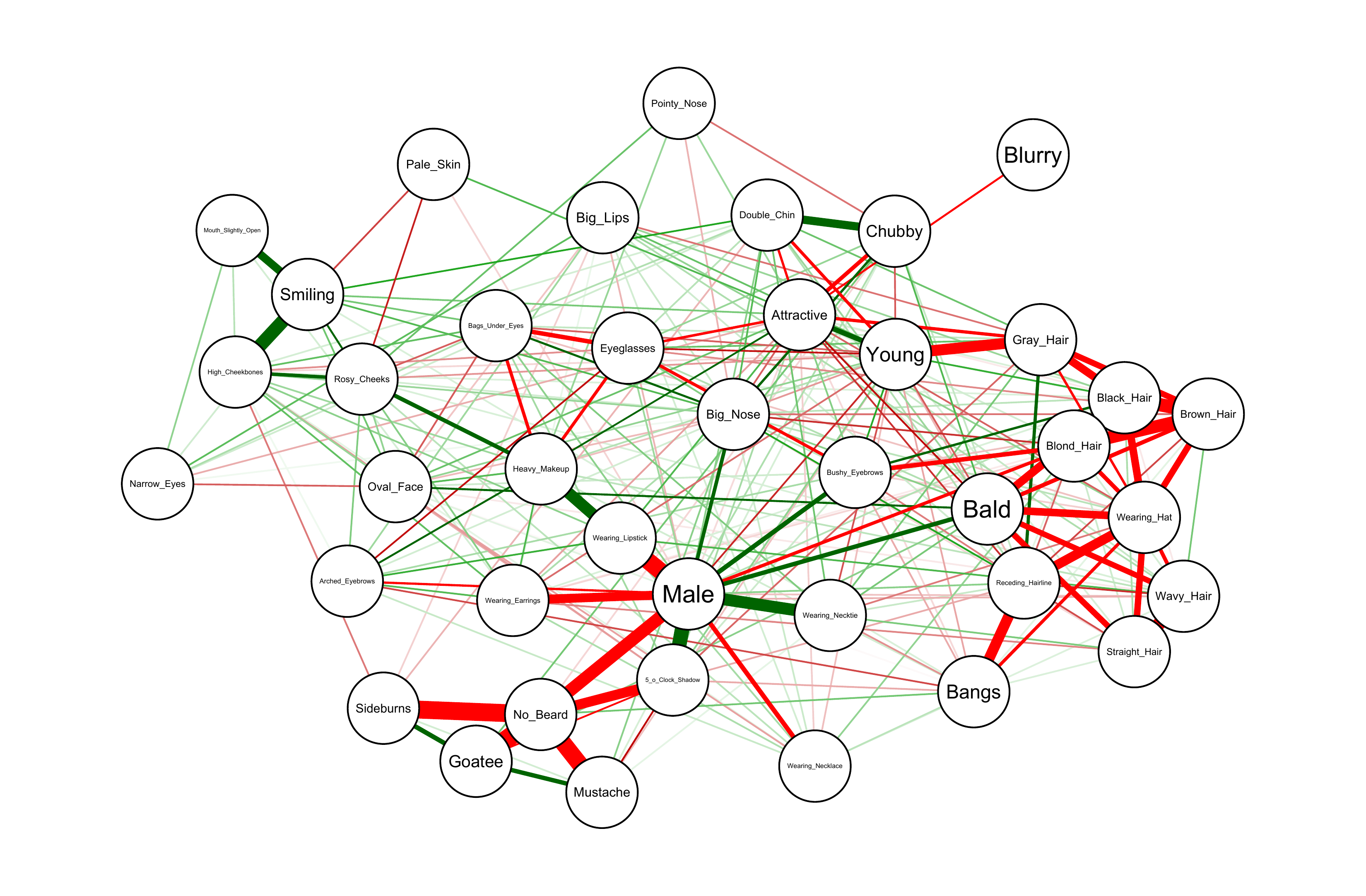}
    \caption{Visualization of Ising model of binary attributes, with green denoting positive edges and red denoting negative edges.}
    \label{fig:G_ising_2}
    \vskip -0.3in
\end{figure*}
\begin{figure}[!htb]
    \begin{center}
    \centerline{\includegraphics[width=0.4\linewidth]{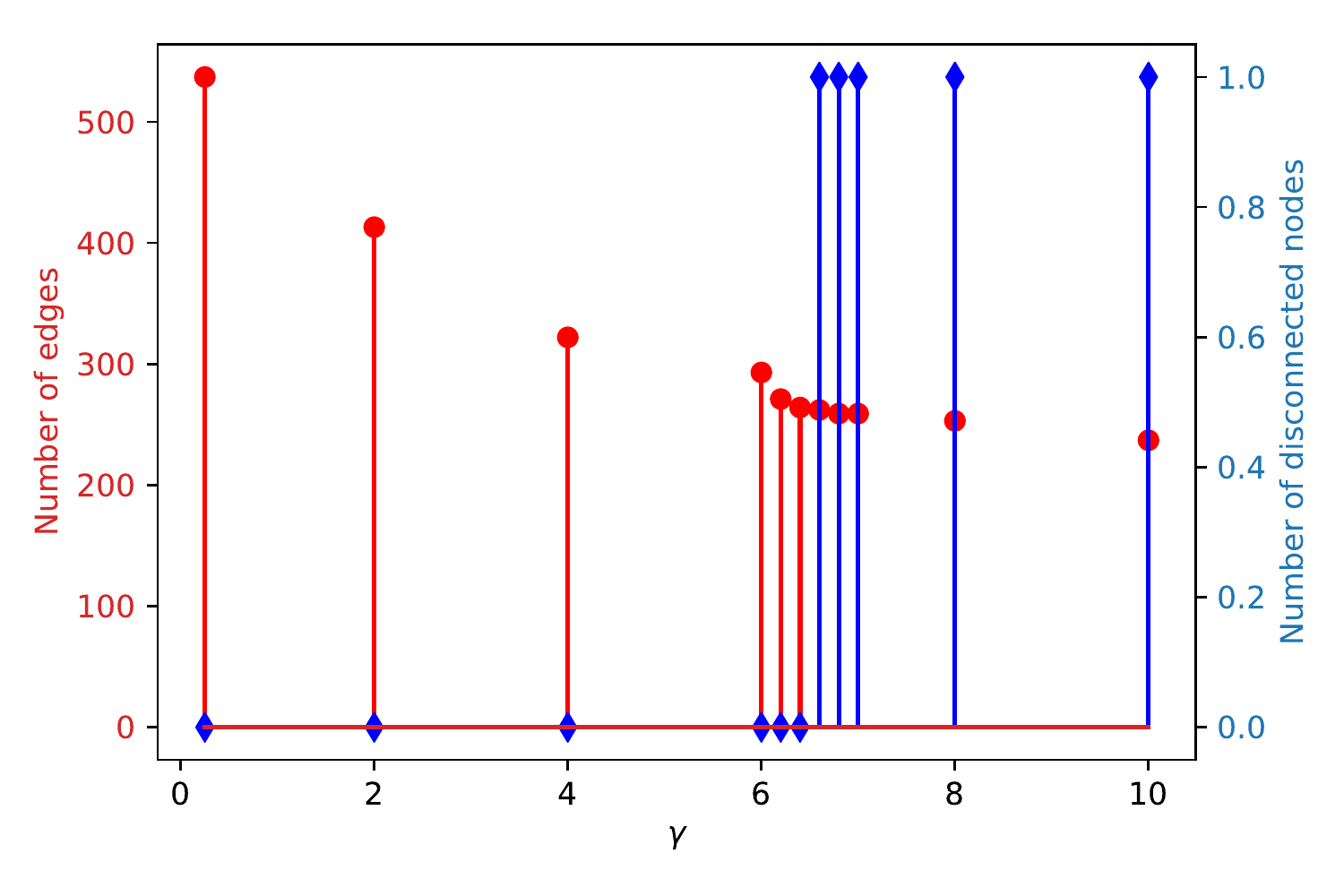}}
    \caption{Goodness-of-fit parameter ($\gamma$) selection.}
    \label{fig:gamma_select}
    \end{center}
\end{figure}
\begin{figure*}[!htb]
    \centering
    \includegraphics[width=\linewidth]{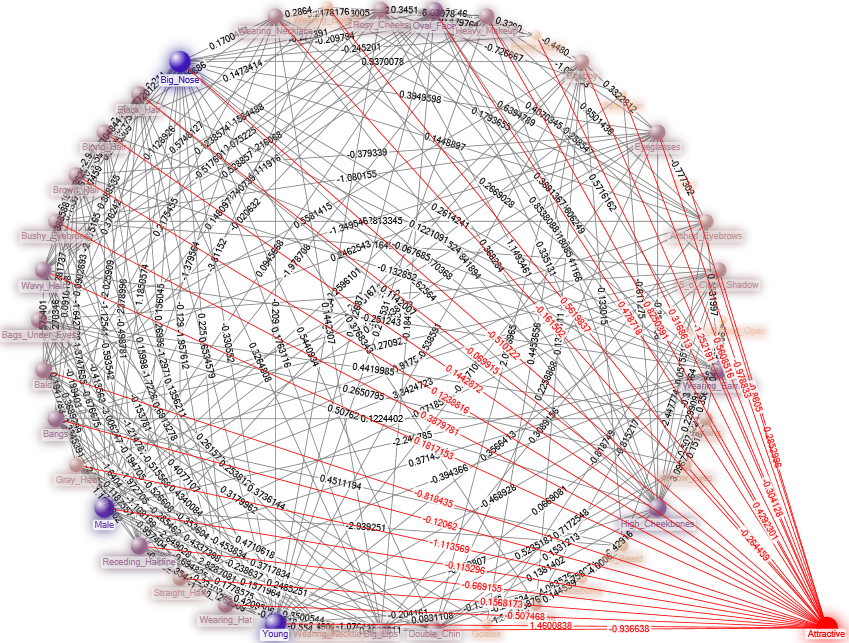}
    \caption{The weighted graph representing the Ising model for the binary attributes of the CelebA-HQ dataset.}
    \label{fig:G_ising}
\end{figure*}

As we learn in the following section, we'd ideally want the learned graph $G^{\ising}(V,E)$ to be connected (no disconnected nodes) as well as sparse. Therefore, we set the the goodness-of-fit parameter ($\gamma$) of the \texttt{eLasso} algorithm to be the largest value above which the resultant graph was disconnected. This variation of the sparsity level of the graph obtained with regards to the number of disconnected nodes as well as the number of edges (a sparsity measure) is seen in Figure~\ref{fig:gamma_select}. As seen, that critical value of the \textit{prior strength} parameter is $\gamma=6.4$.
In Figure~\ref{fig:G_ising} and Figure~\ref{fig:G_ising_2}, we visualize the weighted graph representing the Ising model learned for the $40$ binary attributes of the CelebA-HQ dataset (with $\gamma=6.4$). 

This graph will not just pave the way for the bias corrective procedure to be proposed in the upcoming section, but will also serve as a succinct representation of uncovering anthropocentric bias ingrained during the labeling process itself by the human annotators, by means of Graph exploration using popular GUI Graph visualization software such as Gephi\footnote{\url{https://gephi.org/}} and NodeXL\footnote{\url{https://www.smrfoundation.org/nodexl/}}.

With regards to this \texttt{eLasso} fitting procedure, we have the following figures. In Figure~\ref{fig:G_thresh}, we see the attribute-wise thresholds (or \textit{node parameters} $\{\tau_i\,i \in V\})$. This threshold captures 
the propensity of a specific attribute to be present in an image in the dataset. 

\begin{figure*}[!htb]
    \centering
    \includegraphics[width=\linewidth]{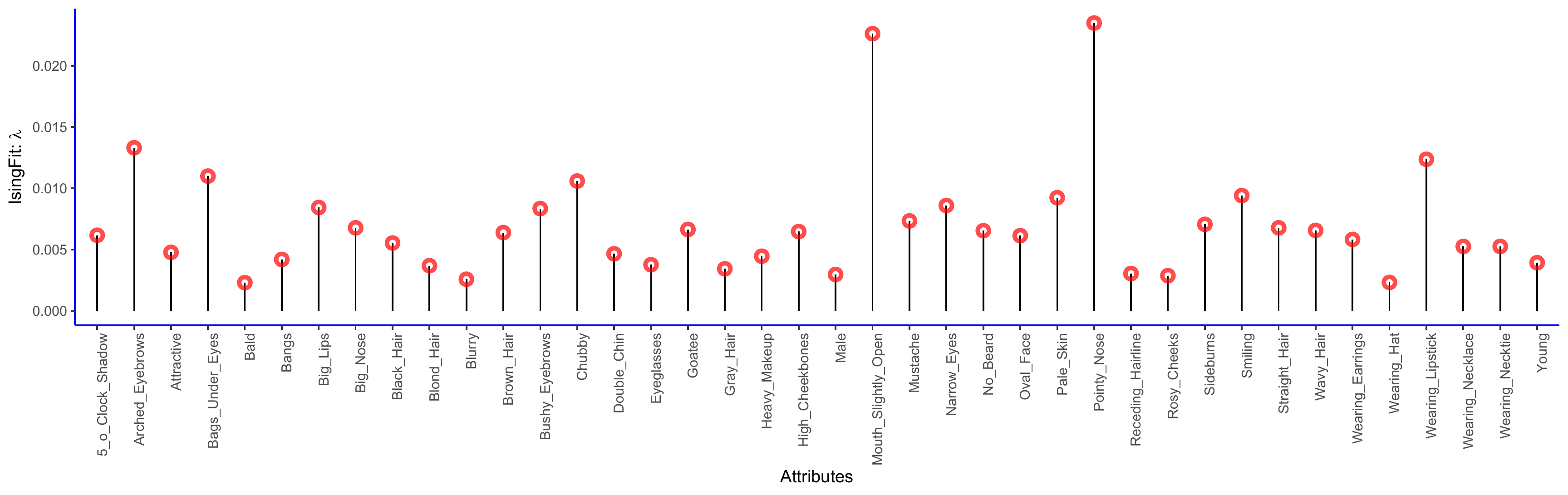}
    \caption{Attribute-wise variation of the tuning parameter ($\lambda$) that was used to achieve the best fit set of neighbors during the Ising prior learning}
    \label{fig:G_lam}
\end{figure*}

As seen, most of the thresholds (with the exception of \texttt{Male}, \texttt{No-Beard} and \texttt{Young}) are all negative. This implies most of the attributes in the CelebA-HQ dataset in-fact exude a normative disposition to be absent.  The attributes \texttt{Rosy-cheeks}, \texttt{Blurry} and \texttt{Wearing-necktie} had the strongest negative thresholds. This implies that these symptoms had the strongest probability of being present in a CelebA-HQ image compared to the other attributes.
In Figure~\ref{fig:G_lam}, we see that variation of the tuning parameter per attribute that was used to ensure the best-fit set of neighbors (see \cite{van2014new}). As seen, the features \texttt{Mouth-slightly-open} and \texttt{Pointy-nose} had the largest values of the tuning parameters associated with them.
\section{Methodology}
\label{methodology}
In this section, we denote the notation $\psi \left( {} \right):{\mathbf{x}} \to {\mathbf{z}}$ denotes the pixel-space to latent-space \textit{encoder} and $\phi \left( {} \right):{\mathbf{z}} \to {\mathbf{x}}$ denotes the latent-space to pixel-space \textit{decoder}. We begin by reviewing the current procedure\footnote{\url{https://github.com/openai/glow/tree/master/demo}} for latent space manipulation and then propose our correction.
\subsection{Current procedure for latent space manipulation}
The current procedure for latent space interpolation firsts  encode inputs and compute the average latent vector of inputs with and without the attribute. The latent space manipulation vectors are 
\begin{equation}\label{eq:z_naive}
    {{\mathbf{Z}}^{(manip)}} = \left\{ {{\mathbf{z}}_i^{(manip)}} \right\}_{i = 1}^{40}
\end{equation}
where for the $i^{th}$ attribute, 
\begin{equation}
{\mathbf{z}}_i^{(manip)} = \frac{{\sum\limits_{t = 1}^{{N_{\dataset}}} {\psi \left( {{{\mathbf{x}}_t}\cdot \mathbbm{1}\left[\kern-0.15em\left[ {{A_{ti}} =  =  + 1} 
 \right]\kern-0.15em\right]} \right)} }}{{\sum\limits_{t = 1}^{{N_{\dataset}}} {\mathbbm{1}\left[\kern-0.15em\left[ {{A_{ti}} =  =  + 1} 
 \right]\kern-0.15em\right]} }} - \frac{{\sum\limits_{t = 1}^{{N_{\dataset}}} {\psi \left( {{{\mathbf{x}}_t}\cdot \mathbbm{1}\left[\kern-0.15em\left[ {{A_{ti}} =  =  0} 
 \right]\kern-0.15em\right]} \right)} }}{{\sum\limits_{t = 1}^{{N_{\dataset}}} {\mathbbm{1}\left[\kern-0.15em\left[ {{A_{ti}} =  =  0} 
 \right]\kern-0.15em\right]} }}
 \label{z_i_orig}
 \end{equation}
 Here, $A\in \{0,1\}^{N_{\dataset} \times N_{\attributes}}$ represents the binary attribute matrix for the dataset and $A_{ti}$ represents the binary value for the $i^{th}$ attribute in the $t^{th}$ image out of $N_{\dataset}$ images in the dataset.
 $\mathbbm{1}\left[\kern-0.15em\left[ {e} 
 \right]\kern-0.15em\right]$ represents the identity function \begin{equation}
     \mathbbm{1}\left[\kern-0.15em\left[ e 
\right]\kern-0.15em\right] = \left\{ {\begin{array}{*{20}{c}}
  1, &\text{if e = TRUE} \\ 
  0, &\text{otherwise.}
\end{array}} \right.
 \end{equation}
After obtaining the latent space manipulation vectors, we can manipulate over the latent space given an input tuple of image, attribute-index and weight of manipulation $(\mathbf{x},i,\alpha )$. By first encoding a raw image $\mathbf{x}$ in latent space as $\psi(\mathbf{x})$, One can interpolate over the latent space by some degree $\alpha$:
\begin{equation}
{{\mathbf{z}}^{\left( {manip\_orig,i,\alpha } \right)}} = \psi \left( {\mathbf{x}} \right) + \alpha {\mathbf{z}}_i^{(manip)}
\end{equation}
To obtain the manipulated image, the interpolated latent vector is decoded to obtain ${{{\mathbf{z}}^{\left( {manip\_orig,i,\alpha } \right)}}}$:
\begin{equation}
{{\mathbf{x}}^{\left( {manip\_orig,i,\alpha } \right)}} = \phi \left( {{{\mathbf{z}}^{\left( {manip\_orig,i,\alpha } \right)}}} \right)
\end{equation}

\subsection{Proposed modified procedure for latent space manipulation}
We suggest that we first realign the latent space attribute-wise manipulation vectors by adding a \textit{Markov-blanket} correction factor to the vectors computed in Eq \ref{z_i_orig} that takes into account the \textit{neighboring} attributes (nodes) unearthed in the Ising prior graph learned from the dataset in Section 2. That is,
we generate different latent manipulation vectors derived from Equation \ref{eq:z_naive} as
\begin{equation}
    {{\mathbf{Z}}^{(manip\_mb)}} = \left\{ {{\mathbf{z}}_i^{(manip\_mb)}} \right\}_{i = 1}^{40}
\end{equation}
where for the $i^{th}$ attribute,
\begin{equation}
{\mathbf{z}}_i^{(manip\_mb)} = {\mathbf{z}}_i^{(manip)}\overbrace { - \mu \sum\limits_{k \in {\mathcal{N}_i}\left( {G^{\ising}(V,E)} \right)} {\left( {{w_{ik}}{\mathbf{z}}_k^{(manip)}} \right)} }^{markov - blanket - correction}.
\label{z_i_mb}
\end{equation}
Here $\mu$ is the Markov-blanket correction strength parameter\footnote{In our experiments, we empirically found that varying $\mu$ between 0.1 to 0.2 yielded the most aesthetically pleasing results.}, ${\mathcal{N}_i}$ is the Markov-blanket neighborhood of the $i^{th}$ attribute (node) in the Ising-graph ${G^{\ising}(V,E)}$ learned in Section~\ref{learning-ising-prior}.

Now, the output images with Markov-blanket corrections, ${{\mathbf{x}}^{\left( {manip\_mb,i,\alpha } \right)}}$, are generated by:
\begin{equation}
\begin{gathered}
  {{\mathbf{z}}^{\left( {manip\_mb,i,\alpha } \right)}} = \psi \left( {\mathbf{x}} \right) + \alpha \left( {{\mathbf{z}}_i^{(manip)}\overbrace { - \mu \sum\limits_{k \in {N_i}\left( {{G^{({\text{ising}})}}(V,E)} \right)} {\left( {{w_{ik}}{\mathbf{z}}_k^{(manip)}} \right)} }^{markov - blanket - correction = {\mathbf{z}}_i^{(manip\_mb)}}} \right) \hfill \\
  {{\mathbf{x}}^{\left( {manip\_mb,i,\alpha } \right)}} = \phi \left( {{{\mathbf{z}}^{\left( {manip\_mb,i,\alpha } \right)}}} \right) \hfill \\ 
\end{gathered} 
\label{x_mb}
\end{equation}
Given the proposed modified method of correctly manipulating across the latent space, we define the full algorithm as per Algorithm \ref{alg:manip}.
\begin{algorithm}[h]
  \caption{Corrective Bias Latent Manipulation Algorithm}
  \label{alg:manip}
\begin{algorithmic}
  \STATE {\bfseries Input:} data $\bm{x}$, priors $\bm{A}$, attributes to interpolate $A$, \\
  degree of interpolation $\alpha$, hyperparameter $\gamma, \mu \in \mathbb{R}^+$
  \STATE $G(V, E) \leftarrow$  Ising Graph of ($\bm{A}, \gamma$)
  \STATE $\psi \leftarrow$ model mapping $\bm{x}$ to $\bm{z}$
  \STATE $\phi \leftarrow$ model mapping $\bm{z}$ to $\bm{x}$
  \STATE $z_m \leftarrow$ manipulation vectors 
  \STATE Initialize $\bm{z'}$ as $\psi(\bm{x})$
  \FOR{$\alpha_i = \alpha_1, \alpha_2, \dots, \alpha_m$ {\bfseries in} $\bm{\alpha}$}
  \STATE $\bm{z'} \leftarrow \bm{z'} + \alpha z_{m_i}$
  \STATE $\mathcal{N} \leftarrow$ neighborhood/adjacent vertices of $z_{m_i}$ in $G$
  \FOR{$k$ {\bfseries in} $\mathcal{N}$}
  \STATE $\bm{z'} \leftarrow \bm{z'} + -\alpha \cdot \mu \cdot w_{ik}(z_{m_k})$
  \ENDFOR
  \ENDFOR
  \STATE $\bm{\hat{x}} \leftarrow \phi(\bm{z'})$, the interpolated image with corrective bias
\end{algorithmic}
\end{algorithm}

\section{Results and Discussion}
\label{results}
In this section, we focus on showcasing the efficacy of our procedure by qualitative methods. Before we dive into the qualitative experiments, we'd like explore if the (Ising) weights of $G^{\ising}(V,E)$, that is, $\left\{ {{w_{ij}}} \right\};\forall (i,j) \in E$ that are indeed correlated with the respective (angular) distance between the latent vectors $\mathbf{z}_i,\mathbf{z}_j$. This would help broadly understand how much of the separation between the latent space representations is on account of mere co-occurrence of the attributes and how much can be attributed to the deep-encoder being able to organically tease out the semantic separation between the attributes.
\begin{figure}[!htb]
    \centering
    \includegraphics[width=0.4\linewidth]{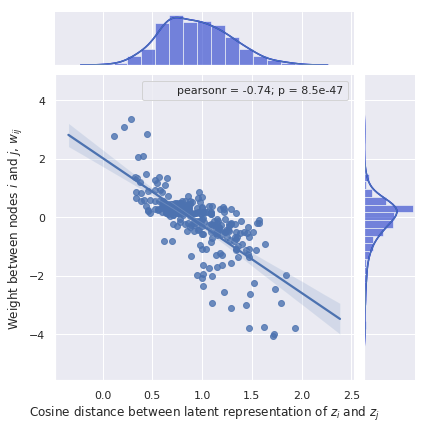}
    \caption{Plot of Ising weights against latent manipulation vectors between the connected nodes in Ising graph.}
    \label{fig:scatter}
\end{figure}

Figure~\ref{fig:scatter} presents the scatter-plots between the Ising weights $w_{ij}$ and $d_{ij}$ as the cosine distance between $\mathbf{z}_i$ and $\mathbf{z}_j$, the latent manipulation vectors associated with attribute indices $i$ and $j$.

The strong negative correlation $(R_{\mathit{pearson}}=-0.74;p=8.5e^{-47})$ confirms our hunch that the co-occurrence of attributes has strongly influenced the latent space. That is, two attributes that always co-occurred in the image dataset (meaning high edge-weight $w_{ij}$ in the Ising prior) ended up with latent space representations with a low cosine distance between them. We posit this might be more of a bug than a feature of the model as this alludes towards the notion that coincidental occurrences will strongly influence the semantic distancing in the latent space. %

\begin{figure}[!htb]
    \centering
    \includegraphics[width=0.8\linewidth]{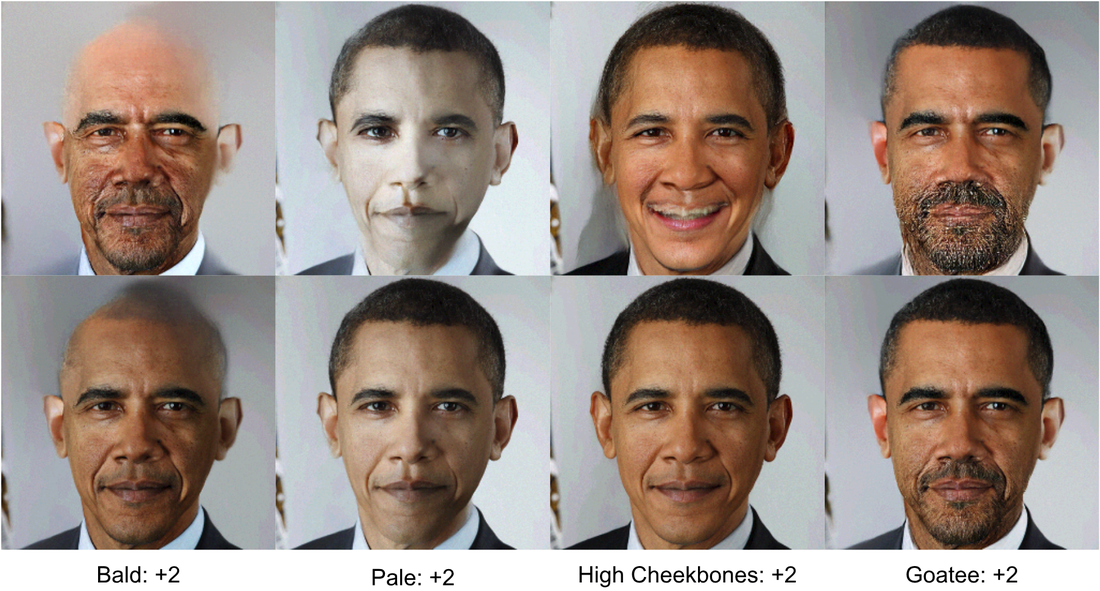}
    \caption{Top: Naively manipulating attribute by +2. Bottom: manipulating attribute by +2 while applying corrective bias penalty to corresponding Markovian blankets.}
    \label{fig:correction}
\end{figure}

\begin{figure}[!htb]
    \centering
    \includegraphics[width=0.8\linewidth]{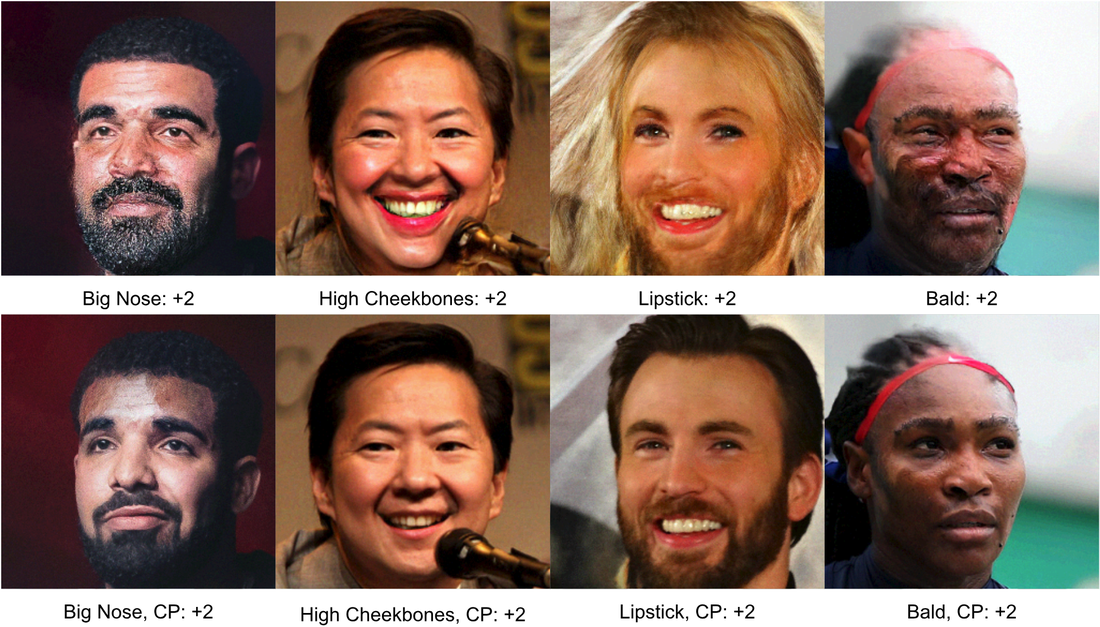}
    \caption{Naive manipulation across one attribute (top) vs manipulation with corrective penalty on neighborhood attributes.}
    \label{fig:two_rows}
\end{figure}

\begin{figure*}[!htb]
    \centering
    \includegraphics[width=\linewidth]{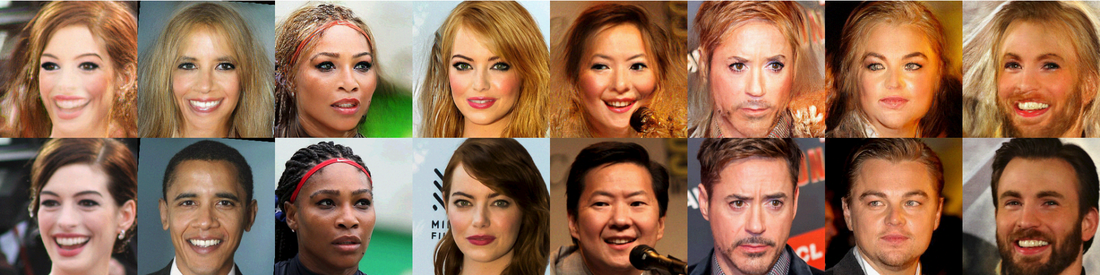}
    \caption{Demonstration on how inherent bias in the 'wearing lipstick' attribute commonly correlated with blonde hair, young and women can be corrected using our proposed corrective bias algorithm. \textit{Top: wearing lipstick: +2. Bottom: wearing lipstick: +2  with neighborhood corrective penalty.}}
    \label{fig:lipstick}
\end{figure*}

Some attributes are mutually associated with others through stereotypes, which experimentally justifies the need for bias correction. As shown in Figure~\ref{fig:two_rows}, increasing \texttt{High\_Cheekbones} also subtly increases the attributes of \texttt{Smiling}, \texttt{Female} and \texttt{Wearing\_Lipstick} which are usually associated with feminine characteristics. On the other hand, increasing attributes such as \texttt{Bald} also shows increases likelihood of \texttt{Goatee}, \texttt{Chubby} and \texttt{Male}. Applying such attributes in images in contrasting domains (increasing \texttt{Lipstick} for male, \texttt{Bald} for females) cause the transformed images to lose intrinsic properties of the original image by generalizing towards the implicitly learnt bias. This is also confirmed in Figure~\ref{fig:correction}, where increasing baldness changes the skin color to be that of a white forehead.

We also demonstrate how biases can be corrected as shown in Figure~\ref{fig:lipstick}. Using \texttt{Wearing\_Lipstick} as an example, which is implicitly associated with feminine characteristics, the image learns to transform with biases such as \texttt{Blonde\_Hair}, \texttt{Female} and \texttt{Wearing\_Makeup}. However, by accounting for biases, we show how \texttt{Wearing\_Lipstick} can be applied across different characteristic domains such as gender and race, while preserving the inherent attributes in the image transformation process.

\section{Conclusion}
We demonstrate that models that directly learn to model the input data without considering attribute priors while training reveal significant bias. Using Glow as an example of flow-based generative models, we reveal the existing bias by manipulating across the latent space, which mirrors common racial and gender stereotypes.

We also emphasize that this bias also exists in other generative models such as GANs and VAEs as shown in the Appendix. We propose a post-hoc corrective measure for models trained with existing bias by using Ising models fitted on attribute priors.

\newpage
\bibliography{example_paper}

\begin{thebibliography}{20}
\providecommand{\natexlab}[1]{#1}
\providecommand{\url}[1]{\texttt{#1}}
\expandafter\ifx\csname urlstyle\endcsname\relax
  \providecommand{\doi}[1]{doi: #1}\else
  \providecommand{\doi}{doi: \begingroup \urlstyle{rm}\Url}\fi

\bibitem[Barocas et~al.()Barocas, Hardt, and Narayanan]{barocasfairness}
Barocas, S., Hardt, M., and Narayanan, A.
\newblock Fairness and machine learning. fairmlbook. org, 2018.
\newblock \emph{URL: http://www. fairmlbook. org}.

\bibitem[Binns(2017)]{binns2017fairness}
Binns, R.
\newblock Fairness in machine learning: Lessons from political philosophy.
\newblock \emph{arXiv preprint arXiv:1712.03586}, 2017.

\bibitem[Bolukbasi et~al.(2016)Bolukbasi, Chang, Zou, Saligrama, and
  Kalai]{bolukbasi2016man}
Bolukbasi, T., Chang, K.-W., Zou, J.~Y., Saligrama, V., and Kalai, A.~T.
\newblock Man is to computer programmer as woman is to homemaker? debiasing
  word embeddings.
\newblock In \emph{Advances in neural information processing systems}, pp.\
  4349--4357, 2016.

\bibitem[Dinh et~al.(2014)Dinh, Krueger, and Bengio]{dinh2014nice}
Dinh, L., Krueger, D., and Bengio, Y.
\newblock Nice: Non-linear independent components estimation.
\newblock \emph{arXiv preprint arXiv:1410.8516}, 2014.

\bibitem[Goodfellow et~al.(2014)Goodfellow, Pouget-Abadie, Mirza, Xu,
  Warde-Farley, Ozair, Courville, and Bengio]{goodfellow2014generative}
Goodfellow, I., Pouget-Abadie, J., Mirza, M., Xu, B., Warde-Farley, D., Ozair,
  S., Courville, A., and Bengio, Y.
\newblock Generative adversarial nets.
\newblock In \emph{Advances in neural information processing systems}, pp.\
  2672--2680, 2014.

\bibitem[He et~al.(2017)He, Zuo, Kan, Shan, and Chen]{he2017attgan}
He, Z., Zuo, W., Kan, M., Shan, S., and Chen, X.
\newblock Arbitrary facial attribute editing: Only change what you want.
\newblock \emph{CoRR}, abs/1711.10678, 2017.
\newblock URL \url{http://arxiv.org/abs/1711.10678}.

\bibitem[Karras et~al.(2017)Karras, Aila, Laine, and
  Lehtinen]{karras2017progressive}
Karras, T., Aila, T., Laine, S., and Lehtinen, J.
\newblock Progressive growing of gans for improved quality, stability, and
  variation.
\newblock \emph{arXiv preprint arXiv:1710.10196}, 2017.

\bibitem[Kingma \& Dhariwal(2018)Kingma and Dhariwal]{kingma2018glow}
Kingma, D.~P. and Dhariwal, P.
\newblock Glow: Generative flow with invertible 1x1 convolutions.
\newblock In \emph{Advances in Neural Information Processing Systems}, pp.\
  10215--10224, 2018.

\bibitem[Kingma \& Welling(2013)Kingma and Welling]{kingma2013auto}
Kingma, D.~P. and Welling, M.
\newblock Auto-encoding variational bayes.
\newblock \emph{arXiv preprint arXiv:1312.6114}, 2013.

\bibitem[Liu et~al.(2019)Liu, Ding, Xia, Liu, Ding, Zuo, and
  Wen]{Liu2019STGANAU}
Liu, M., Ding, Y., Xia, M., Liu, X., Ding, E., Zuo, W., and Wen, S.
\newblock Stgan: A unified selective transfer network for arbitrary image
  attribute editing.
\newblock 2019.

\bibitem[Liu et~al.(2015)Liu, Luo, Wang, and Tang]{liu2015faceattributes}
Liu, Z., Luo, P., Wang, X., and Tang, X.
\newblock Deep learning face attributes in the wild.
\newblock In \emph{Proceedings of International Conference on Computer Vision
  (ICCV)}, 2015.

\bibitem[Sainburg et~al.(2018)Sainburg, Thielk, Theilman, Migliori, and
  Gentner]{sainburg2018generative}
Sainburg, T., Thielk, M., Theilman, B., Migliori, B., and Gentner, T.
\newblock Generative adversarial interpolative autoencoding: adversarial
  training on latent space interpolations encourage convex latent
  distributions.
\newblock \emph{arXiv preprint arXiv:1807.06650}, 2018.

\bibitem[Stanley(1971)]{stanley1971phase}
Stanley, H.~E.
\newblock \emph{Phase transitions and critical phenomena}.
\newblock Clarendon Press, Oxford, 1971.

\bibitem[Tom~Yeh et~al.(2017)]{tom2017designing}
Tom~Yeh, M. et~al.
\newblock Designing a moral compass for the future of computer vision using
  speculative analysis.
\newblock In \emph{Proceedings of the IEEE Conference on Computer Vision and
  Pattern Recognition Workshops}, pp.\  64--73, 2017.

\bibitem[Tommasi et~al.(2017)Tommasi, Patricia, Caputo, and
  Tuytelaars]{tommasi2017deeper}
Tommasi, T., Patricia, N., Caputo, B., and Tuytelaars, T.
\newblock A deeper look at dataset bias.
\newblock In \emph{Domain adaptation in computer vision applications}, pp.\
  37--55. Springer, 2017.

\bibitem[Torralba et~al.()Torralba, Efros, et~al.]{torralba2011unbiased}
Torralba, A., Efros, A.~A., et~al.
\newblock Unbiased look at dataset bias.
\newblock Citeseer.

\bibitem[Van~Borkulo et~al.(2014)Van~Borkulo, Borsboom, Epskamp, Blanken,
  Boschloo, Schoevers, and Waldorp]{van2014new}
Van~Borkulo, C.~D., Borsboom, D., Epskamp, S., Blanken, T.~F., Boschloo, L.,
  Schoevers, R.~A., and Waldorp, L.~J.
\newblock A new method for constructing networks from binary data.
\newblock \emph{Scientific reports}, 4:\penalty0 5918, 2014.

\bibitem[Wainwright et~al.(2008)Wainwright, Jordan,
  et~al.]{wainwright2008graphical}
Wainwright, M.~J., Jordan, M.~I., et~al.
\newblock Graphical models, exponential families, and variational inference.
\newblock \emph{Foundations and Trends{\textregistered} in Machine Learning},
  1\penalty0 (1--2):\penalty0 1--305, 2008.

\bibitem[Xiao et~al.(2018)Xiao, Hong, and Ma]{xiao2018dna}
Xiao, T., Hong, J., and Ma, J.
\newblock Dna-gan: Learning disentangled representations from multi-attribute
  images.
\newblock \emph{International Conference on Learning Representations,
  Workshop}, 2018.

\bibitem[Zhou et~al.(2017)Zhou, Xiao, Yang, Feng, He, and He]{zhou2017genegan}
Zhou, S., Xiao, T., Yang, Y., Feng, D., He, Q., and He, W.
\newblock Genegan: Learning object transfiguration and attribute subspace from
  unpaired data.
\newblock \emph{arXiv preprint arXiv:1705.04932}, 2017.

\end{thebibliography}
\bibliographystyle{icml2019}

\newpage

\section{Appendix}
\subsection{Visializing Learnt Biases in Generative Models}
\begin{figure*}[!htb]
    \centering
    \vskip 0.1in
    \includegraphics[width=\linewidth]{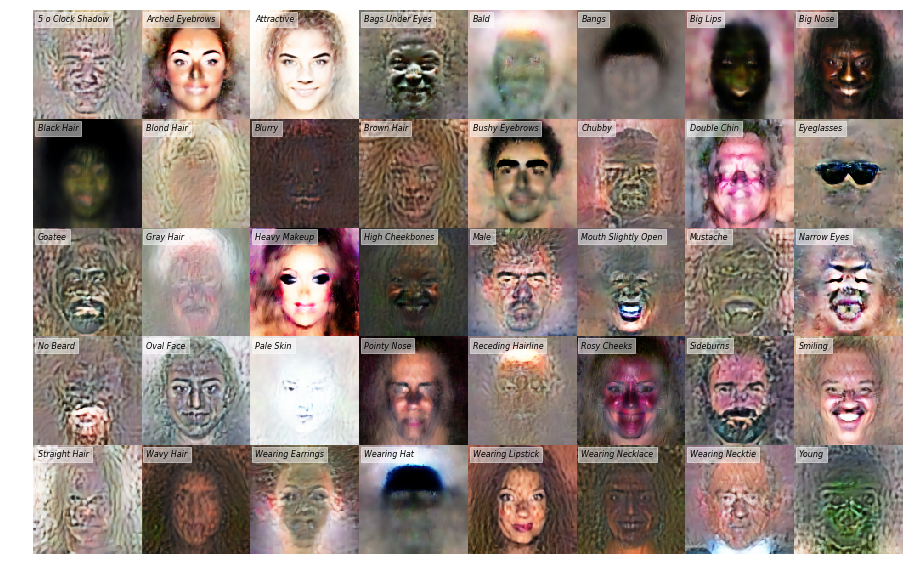}
    \vskip -0.1in
    \caption{Bias of CelebA illustrated: Visualization of biases in latent space of Generative Adversarial Interpolative Autoencoding (GAIA) \cite{sainburg2018generative}. Visualization of latent `\texttt{attractive}', `\texttt{big nose}', and `\texttt{big lips}' correlate to racial profiles that reveal the inherent bias in CelebA.}
    \label{fig:gaia}
\end{figure*}
\subsection{Visualization of Ising Graphs}

\begin{figure*}[!htb]
    \centering
    \includegraphics[width=\linewidth]{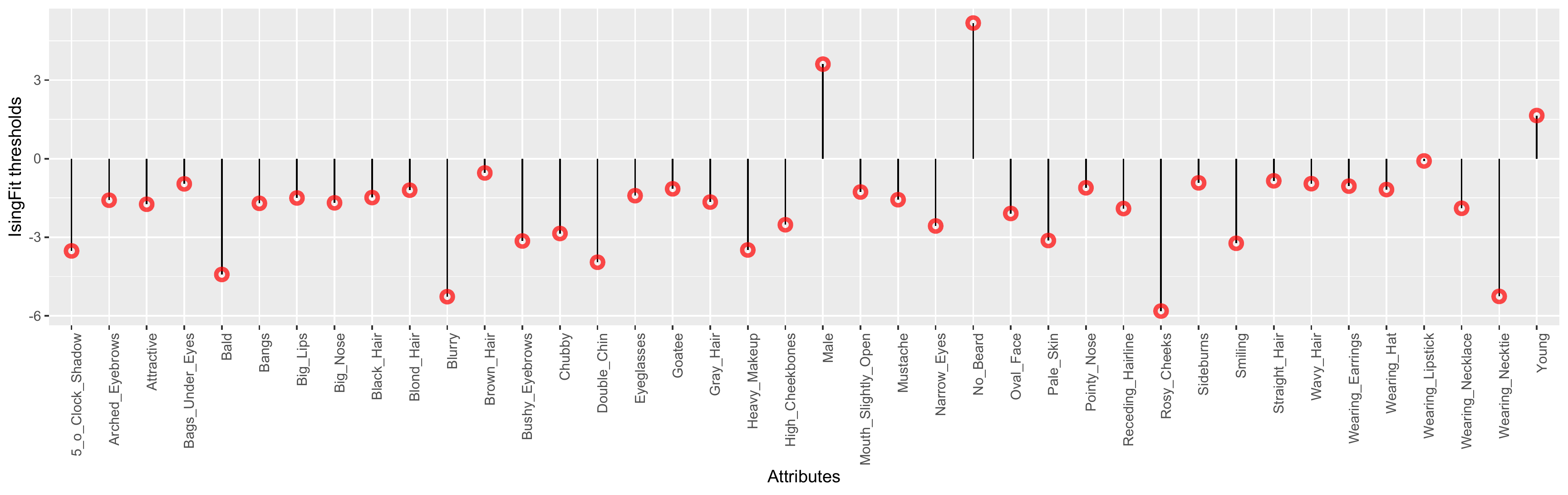}
    \caption{Attribute-wise thresholds (or \textit{node parameters} $\tau_i$) used during the \texttt{eLASSO} fitting procedure}
    \label{fig:G_thresh}
\end{figure*}

\subsection{Generalizing beyond Normalizing Flows: Revealing Bias in Generative Adversarial Networks Trained on CelebA/CelebA-HQ}
We can further confirm that the CelebA dataset is the root cause of the bias shown experimentally by revealing similar biases in other models, such as Generative Adversarial Networks. Below, we experiment with two GAN architectures trained on the CelebA dataset.

Note that the pretrained models only allow the manipulation of 13 attributes: \texttt{Bald}, \texttt{Bangs}, \texttt{Black\_Hair}, \texttt{Blond\_Hair}, \texttt{Brown\_Hair}, \texttt{Bushy\_Eyebrows}, \texttt{Eyeglasses}, \texttt{Male}, \texttt{Mouth\_Slightly\_Open}, \texttt{Mustache}, \texttt{No\_Beard}, \texttt{Pale\_Skin}, \texttt{Young}. %

\subsubsection{AttGAN: Facial Attribute Editing by Only Changing What You Want}

AttGAN, by He et al., was created specifically to modify attributes realistically and in isolation, so that users can ``change what [they] want" \cite{he2017attgan}. Users can choose to modify any number of attributes simultaneously, as well as adjust the magnitude of adjustment per attribute.

The authors of AttGAN understand that attributes in CelebA are highly correlated, claiming that in previous models, ``adding blond hair
always makes a male become a female because most blond
hair objects are female in the training set". He et al.\ correct the effects of this correlation by placing constraints on the images generated from latent representations, enforcing these constraints by employing an attribute classifier, a main contribution of their paper.

We begin by observing that adjusting the \texttt{Male} attribute negatively causes the AttGAN model to generate long hair and add makeup. Next, we adjust the \texttt{Blond\_Hair} attribute, and observe that this causes the GAN to generate blue eyes and long hair.

\begin{figure}[H]
    \centering
    \includegraphics[width=0.7\textwidth]{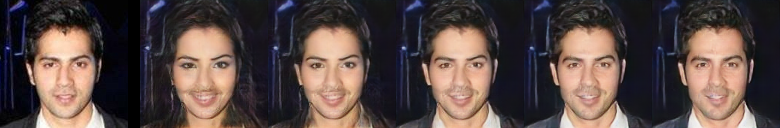}
    \includegraphics[width=0.7\textwidth]{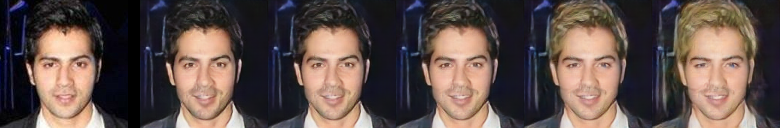}
    \caption{Top: Example of interpolating \texttt{Male} attribute between -2 and +2. Bottom: Example of interpolating \texttt{Blond\_Hair} attribute between -2 and +2. Leftmost image is original image.}
\end{figure}

We also increase the \texttt{Blond\_Hair} attribute and the \texttt{Male} attributes together. One might expect increasing the \texttt{Male} attribute to counteract the long-hair that the \texttt{Blond\_Hair} attribute creates. However, as we can see below, it on occasion acts constructively. It is clear that, even with the addition of an attribute classifier to constrain the model, the latent representation manipulations for separate attributes are still not independent. We note, though, that this behavior is seen mostly when the original images are of males. He et al. themselves conclude that model errors occur mostly when large modifications are made to images, such as adding hair to a bald figure.

Lastly, we investigate the manipulation of facial hair. It seems that mustaches and beards are always generated black regardless of the original subject's hair color. In fact, if we attempt to generate a mustache on top of a subject who already has facial hair, it simply changes the color of the existing facial hair to black. Additionally, if we generate a mustache on top of a young or less masculine subject, the rest of the face is morphed into that of a more elderly and rugged man.

\begin{figure}[H]
    \centering
    \includegraphics[width=0.23\textwidth]{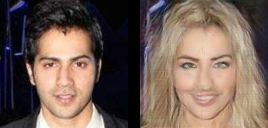}
    \includegraphics[width=0.23\textwidth]{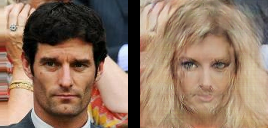}\\
    \includegraphics[width=0.23\textwidth]{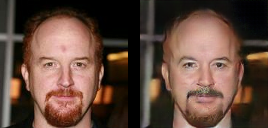}
    \includegraphics[width=0.23\textwidth]{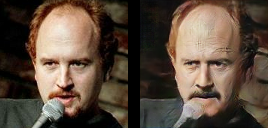}\\
    \includegraphics[width=0.23\textwidth]{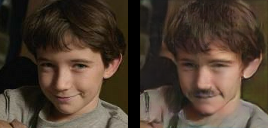}
    \includegraphics[width=0.23\textwidth]{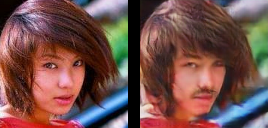}
    \caption{Top: Examples of manipulating both \texttt{Male} and \texttt{Blond\_Hair} attributes by +1.\ Middle: Manipulating \texttt{Mustache} attribute by +1 on a subject with existing facial hair. Bottom: Manipulating \texttt{Mustache} attribute by +1 on a young subject with less masculine features.}
\end{figure}

\subsubsection{STGAN: A Unified Selective Transfer Network for Arbitrary Image Attribute Editing}
We also carry out experiments on STGAN, a model built by Liu et al.\ \cite{Liu2019STGANAU} extending He et al.'s\ work. Their main contribution is the selectivity of information passed into their model's generators, choosing only to pass in the difference in attribute vector between the source and target image, so that the generator supposedly only affects the targeted change in attribute.

As with AttGAN, we begin by manipulating the \texttt{Male} attribute. We can see that STGAN is less inclined to generate longer hair when decreasing this attribute, but it generates black splotches in the shape of hair, simply without the texture and details of hair. It is obvious that STGAN still suffers from the safe bias in the priors as other models.

Like AttGAN, STGAN changes the eye color of its subject towards blue when increasing the \texttt{Blond\_Hair} attribute. However, it does not generate long hair like AttGAN does.

\begin{figure}[H]
    \centering
    \includegraphics[width=0.7\textwidth]{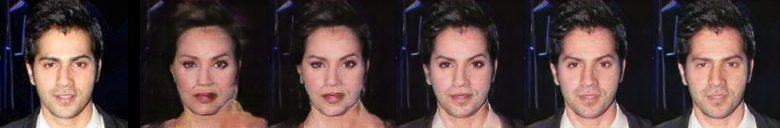}
    \includegraphics[width=0.7\textwidth]{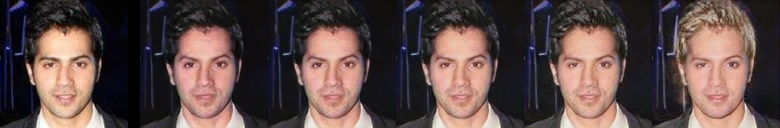}
    \caption{Top: Example of interpolating \texttt{Male} attribute between -2 and +2. Bottom: Example of interpolating \texttt{Blond\_Hair} attribute between -2 and +2. Leftmost image is original image.}
\end{figure}

Increasing the \texttt{Blond\_Hair} and the \texttt{Male} attributes together reproduces the constructive interaction seen in AttGAN, where even more blond long hair is generated than increasing just \texttt{Blond\_Hair} on its own. This underlying complex interaction in the latent space seems to be a consistent problem across generative models, as far as we can tell.

Like AttGAN, generated facial hair is always black, regardless of the original subject's hair color, generating a mustache on top of a subject who already has facial hair simply darkens the color of the existing facial hair, and generating a mustache on a young/soft-featured subject makes them appear older and more masculine.

\begin{figure}[H]
    \centering
    \includegraphics[width=0.23\textwidth]{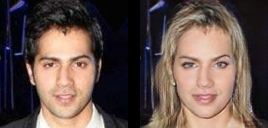}
    \includegraphics[width=0.23\textwidth]{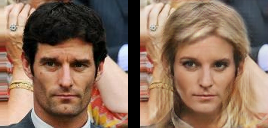}\\
    \includegraphics[width=0.23\textwidth]{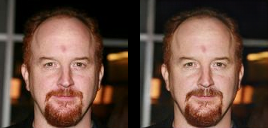}
    \includegraphics[width=0.23\textwidth]{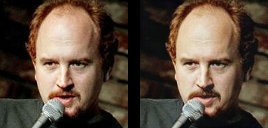}\\
    \includegraphics[width=0.23\textwidth]{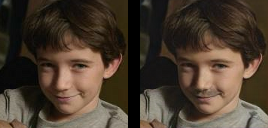}
    \includegraphics[width=0.23\textwidth]{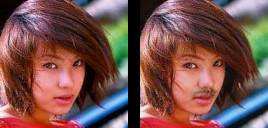}
    \caption{Top: Examples of manipulating both \texttt{Male} and \texttt{Blond\_Hair} attributes by +1. Middle: Manipulating \texttt{Mustache} attribute by +1 on a subject with existing facial hair. Bottom: Manipulating \texttt{Mustache} attribute by +1 on a young subject with less masculine features.}
\end{figure}

\subsubsection{DNA-GAN: Learning Disentangled Representations from Multi-Attribute Images}
\begin{figure}[!htb]
    \centering
    \includegraphics[width=0.7\textwidth]{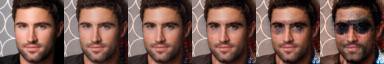}
    \caption{Interpolating on `\texttt{wearing eyeglasses}' changes the racial profile of the subject matter for DNA-GAN.}
    \label{fig:dna}
\end{figure}
DNA-GAN is a supervised method for disentangling multiple factors of variation simultaneously by using multi-attribute images \cite{xiao2018dna}. Trained on CelebA, it can manipulate several attributes in the latent representations of images, which is a generalization of GeneGAN \cite{zhou2017genegan}. DNA-GAN replaces the explicit nulling loss with the annihilating operation and employs a single discriminator for guiding images generation on multiple attributes. 

While it seeks to disentangle multiple factors and attributes of images, we see the inherent bias reflected in the CelebA dataset as per Figure~\ref{fig:dna}. While we manipulate across the `\texttt{wearing eyeglasses}', the racial profile of the subject matter being manipulated also changes accordingly.

\subsubsection{Generative Adversarial Interpolative Autoencoding (GAIA)}

The Generative Adversarial Interpolative Autoencoder (GAIA) is novel hybrid between the Generative Adversarial Network (GAN) and the Autoencoder (AE) \cite{sainburg2018generative}. It addresses the issue of GANs which are non-bidirectional, while also addressing issues of autoencoders which produces blurry images, and addressing the non-conves latent spaces of autoencoders.

GAIA promotes a convex latent distribution by training adversarially on latent space interpolations. Trained on CelebA, GAIA produces non-blurry samples that match both high- and low-level features of the original images. Upon visualizing the inherent latent spaces learnt, we see that attributes such as `\texttt{big nose}' and `\texttt{big lips}' correlate to racial profiles that reveal the inherent bias in CelebA, as per Figure~\ref{fig:gaia}.

\end{document}